\documentclass[journal]{IEEEtran}
\usepackage{setspace}
\usepackage{amsmath}
\usepackage{amsfonts}
\usepackage{indentfirst}
\usepackage{cite}
\usepackage{bm}
\usepackage{graphicx}
\usepackage{subfigure}
\usepackage{booktabs}
\usepackage{multirow}

\begin{document}

\title{Efficient Outlier Removal in Large-scale Global Structure-from-Motion}

\author{Fei Wen, 
        Danping Zou,
        Rendong Ying,
        and Peilin Liu

\thanks{F. Wen, D. Zou, R. Ying and P. Liu are with the Department of Electronic Engineering, Shanghai Jiao Tong University, Shanghai 200240, China (e-mail: wenfei@sjtu.edu.cn; dpzou@sjtu.edu.cn; rdying@sjtu.edu.cn; liupeilin@sjtu.edu.cn).} 
}

\markboth{}
{Shell \MakeLowercase{\textit{et al.}}: Bare Demo of IEEEtran.cls for Journals}

\maketitle

\begin{abstract}
This work addresses the outlier removal problem in large-scale global structure-from-motion. In such applications, outlier removal is very useful to mitigate the deterioration caused by mismatches in the feature point matching step. Unlike existing outlier removal methods, we exploit the structure in multiview geometry problems to propose a dimension reduced formulation, based on which two efficient methods have been developed. The first method considers a convex relaxed $\ell_1$ minimization and is solved by a single linear programming (LP). The second method approximately solves the ideal $\ell_0$ minimization by an iteratively reweighted method. The dimension reduction results in a significant speedup of the new algorithms. Further, the iteratively reweighted method can significantly reduce the possibility of removing true inliers. Results show that, compared with state-of-the-art algorithms (e.g., the $\ell_1$ method), the proposed algorithms are more than three times faster and meanwhile  produce better consensus sets. Matlab code for reproducing the results is available at \textit{https://github.com/FWen/OUTLR.git}.
\end{abstract}


\IEEEpeerreviewmaketitle {}

\section{Introduction}
\label{sec:intro}
Structure-from-Motion (SfM) tries to recover the 3D point clouds and camera poses from a set of unordered images.
There are two approaches to this problem. The first one is the incremental approach. It starts from two images, and grows the point clouds and computes the poses incrementally. The second one is the global approach, which firstly estimates the relative orientations among different images and then recovers both the 3D point positions and camera positions simultaneously. The recovered structures from both approaches are used as initialization in bundle adjustment to get the final results. It has been shown in \cite{28} that for large-scale SfM problems, the global approach works significantly better, since the global approach formulates SfM as a convex optimization problem, which guarantees the global optimum result theoretically.

Though outlier removal has been applied to remove the incorrect point correspondences in the first step of global SfM, which solves the relative orientations from pairwise matching between different views, a large number of outliers still exist because the epipolar constraints among local views usually do not reveal outliers in long point tracks. Those outliers will significantly degrade the structure estimation if they are not well processed [19]. The main interest of this work is to find an efficient and robust solution to address the outlier removal problem in global SfM.



The problem of outlier removal in model fitting, or robust model fitting, is fundamentally critical
to many computer vision applications, including fundamental matrix estimation, homography matrix estimation, vision-based robotics navigation, and global outlier removal in SfM [1]--[3].
Given a set of $M$ measurements ${{\rm{\{ }}{{\bf{a}}_i}{\rm{,}}{y_i}{\rm{\} }}_{i = 1, \cdots M}}$, an important problem arises in many computer vision applications is to remove the outliers in the data set. It is also known as the maximum consensus problem [4], which aims to find a model, parameterized by ${\bf{x}} \in \mathbb{R}{^N}$, that is consistent with as many of the input data as possible, i.e., has the largest consensus set $I$ as [5]
\begin{equation}\label{I-1}
\begin{split}
&\mathop {{\rm{maximize }}}\limits_{{\bf{x}},I \subseteq \Omega } |I|\\
\textrm{subject to  } ~&\left| {{\bf{a}}_i^T{\bf{x}} - {y_i}} \right| \le \delta, ~~\forall i \in I
\end{split}
\end{equation}
where $\delta > 0$ is the inlier threshold, $\Omega = \{ 1,2, \cdots ,M\}$ is the index set.
For a solution ${I^*}$ with size $|{I^*}|$, ${I^*}$ denotes the index set of the true inliers,
and $\Omega \backslash {I^*}$ denotes the index set of the true outliers.

On the one hand, due to the intractability of the robust geometric fitting problem, the global optimum can only be found by searching [25], which makes globally optimal algorithms only suitable for low-dimensional problems. On the other hand, the class of randomized hypothesize-and-verify algorithms are more efficient and popular, e.g., RANSAC [26] and its many variants [27]. Although such randomized algorithms are efficient, they can usually produce only approximate solution and do not guarantee a good estimate due to their randomized nature. Very recently, deterministic methods fill the gap between these two classes have been proposed in [4], [24]. Such deterministic methods are much more efficient than globally optimal algorithms, while being able to achieve better solution quality compared with hypothesize-and-verify based randomized algorithms.

Due to the high-dimensionality of the global outlier removal problem in global SfM, the methods in [4], [5], [24]--[27] are usually impractical and not applicable. The most efficient algorithm suitable for this problem is the ${\ell _1}$ method [6], which solves a convex relaxation of (1). Although there is no theoretical guarantee of success, the ${\ell _1}$ method works very well in practice and can yield a significant drop in reprojection error.

In this work, we propose outlier removal algorithms that are more efficient than
the ${\ell _1}$ method [6] and suitable for large-scale Global SfM. The key idea is that
we use a dimension reduced formulation to reduce the computational complexity. Similar to [6], we present two
versions of outlier removal algorithms. The first fast version formulates the outlier removal as
a convex relaxed ${\ell _1}$ minimization that is solved by linear programming (LP).
The second one approximately solves the ideal ${\ell _0}$ formulation by an iteratively reweighted algorithm with only a little bit of efficiency trade-off.

We have conducted experiments on both synthetic and real world datasets.
Experimental results show that our methods retain more inliers than existing methods, and run much faster.

\section{Proposed Formulation and Connection to Existing Works}

\subsection{Proposed Formulation}

Using a nonnegative auxiliary variable ${\bf{s}} \in \mathbb{R}{^M}$, ${\bf{s}} \ge 0$,
problem (1) can be recast into the following formulation
\begin{equation}\label{I-2}
\begin{split}
&\mathop {{\rm{maximize }}}\limits_{{\bf{x}},{\bf{s}}} M - {\left\| {\bf{s}} \right\|_0}\\
\textrm{subject to  } ~&\left| {{\bf{a}}_i^T{\bf{x}} - {y_i}} \right| \le \delta  + {s_i}, ~~{s_i} \ge 0
\end{split}
\end{equation}
where ${\left\| {\bf{s}} \right\|_0}$ denotes the ${\ell _0}$ norm which counts the number of nonzero elements in ${\bf{s}}$. For a solution ${{\bf{s}}^*}$ of (\ref{I-2}), it holds ${\rm{supp}}({{\bf{s}}^*}) = \Omega \backslash {I^*}$, which is the index set of the outliers, and $M - {\left\| {\bf{s}} \right\|_0} = |{I^*}|$.
${\rm{supp}}( \cdot )$ denotes the support set (the index set of nonzero) of a vector.
Equivalently, problem (\ref{I-2}) can be expressed as the constrained ${\ell _0}$ minimization
\begin{equation}\label{I-3}
\begin{split}
&\mathop {{\rm{minimize }}}\limits_{{\bf{x}},{\bf{s}}} {\left\| {\bf{s}} \right\|_0}\\
\textrm{subject to  } ~&\left| {{\bf{a}}_i^T{\bf{x}} - {y_i}} \right| \le \delta  + {s_i},~~ {s_i} \ge 0.
\end{split}
\end{equation}

The constraint in (\ref{I-3}) can be expressed as linear inequalities.
Specifically, the inequality constraint in (\ref{I-3}) is equivalent to
\begin{equation}\label{I-4}
{\bf{a}}_i^T{\bf{x}} - {y_i} \le \delta  + {s_i}~~~{\rm{and}}~~~- {\bf{a}}_i^T{\bf{x}} + {y_i} \le \delta  + {s_i}.
\end{equation}
Let ${\bf{A}} = {\left[ {{\bf{A}}_1^T,{\bf{A}}_2^T, \cdots ,{\bf{A}}_M^T} \right]^T}$
and ${\bf{b}} = {\left[ {{\bf{b}}_1^T,{\bf{b}}_{\rm{2}}^T, \cdots ,{\bf{b}}_M^T} \right]^T}$ with
${{\bf{A}}_i} = {\left[ {{{\bf{a}}_i}, - {{\bf{a}}_i}} \right]^T}$
and ${{\bf{b}}_i} = {\left[ {{y_i} + \delta , - {y_i} + \delta } \right]^T}$.
Then, the formulation (\ref{I-3}) can be rewritten as
\begin{equation}\label{I-5}
\begin{split}
&\mathop {{\rm{minimize }}}\limits_{{\bf{x}},{\bf{s}}} {\left\| {\bf{s}} \right\|_0}\\
\textrm{subject to  } ~&{\bf{Ax}} \le {\bf{b}} + {\bf{s}} \otimes {{\bf{1}}_{\kappa  \times 1}},~~{\bf{s}} \ge 0
\end{split}
\end{equation}
where $\kappa = 2$, ${\bf{1}}_{\kappa  \times 1}$ is a $\kappa$-dimensional vector with all elements be 1, and $\otimes$ denotes the Kronecker product.

\subsection{Multiview Geometry}

The above formulations are designed for the linear regression residual $\left| {{\bf{a}}_{}^T{\bf{x}} - y} \right|$, which can be extended to handle geometric residuals, e.g., in multiview reconstruction. In multiview geometry, the goal is to estimate the structure of the scene and the camera motion from image projections. For instance, let $(z_1^i,z_2^i)$ be a measurement in one of the images and ${\bf{z}}$ be its corresponding 3D-point. Given the camera matrix ${\bf{P}} = [{\bf{R}},{\bf{t}}] \in \mathbb{R}{^{3 \times 4}}$, the squared reprojection error is [6]
\begin{equation}\label{I-6}
{E_i}({\bf{z}},{\bf{P}}) = \left\| {z_1^i - \frac{{{{\bf{r}}_1}{\bf{z}} + {t_1}}}{{{{\bf{r}}_3}{\bf{z}} + {t_3}}},z_2^i - \frac{{{{\bf{r}}_2}{\bf{z}} + {t_2}}}{{{{\bf{r}}_3}{\bf{z}} + {t_3}}}} \right\|_2^2
\end{equation}
where ${{\bf{r}}_j}$ denotes the $j$-th row of ${\bf{R}}$, ${\bf{t}} = {[{t_1},{t_2},{t_3}]^T} \in \mathbb{R}{^3}$ denotes the translation of the camera. In the general framework, a reformulation of the squared error residuals is of the form
\begin{equation}\label{I-7}
{E_i}({\bf{x}}) = \frac{{\left\| {{\bf{u}}_i^T{\bf{x}} + {{\tilde u}_i},{\bf{v}}_i^T{\bf{x}} + {{\tilde v}_i}} \right\|_2^2}}{{{{({\bf{w}}_i^T{\bf{x}} + {{\tilde \omega }_i})}^2}}}.
\end{equation}
In many geometry problems in computer vision, if either $\bf{z}$ or $\bf{R}$ is known, the residual formulation (\ref{I-6}) can be rewritten in the form of (\ref{I-7}). For example, for the triangulation problem, $\bf{R}$ and $\bf{t}$ are kept fixed, and ${\bf{x}}$ denotes the position parameters of 3D-points. For the multiview reconstruction problem, both the positions of the 3D-points and the positions of the cameras are unknown. In this case, $\bf{x}$ in (\ref{I-7}) contains the parameters of the 3D-points and the camera translations, which is also called the known rotation problem [7].

To handle such geometric residuals, in addition to the Euclidian distance, the coordinate-wise max distance (${\ell _\infty }$ norm) and the absolute distance ($\ell_1$ norm) are also popular due to their convenience. Specifically, a general formulation of the error residuals is of the form [7], [8]
\[{E_i}({\bf{x}}) = \frac{{{{\left\| {{\bf{u}}_i^T{\bf{x}} + {{\tilde u}_i},{\bf{v}}_i^T{\bf{x}} + {{\tilde v}_i}} \right\|}_p}}}{{{\bf{w}}_i^T{\bf{x}} + {{\tilde \omega }_i}}}~~{\rm{ with}}~~{\bf{w}}_i^T{\bf{x}} + {\tilde \omega _i} > 0\]
where ${E_i}({\bf{x}})$ is quasi-convex for any $p \ge 1$ [16].
With an inlier threshold $\delta > 0$, the constraint for outlier removal is
\begin{equation}\label{I-8}
{\left\| {{\bf{u}}_i^T{\bf{x}} + {{\tilde u}_i},{\bf{v}}_i^T{\bf{x}} + {{\tilde v}_i}} \right\|_p} \le \delta ({\bf{w}}_i^T{\bf{x}} + {\tilde \omega _i})
\end{equation}
where ${\bf{w}}_i^T{\bf{x}} + {\tilde \omega _i} > 0$ is naturally satisfied as $\delta > 0$.

For the ${\ell _1}$ norm, i.e., $p=1$, the constraint becomes
\[|{\bf{u}}_i^T{\bf{x}} + {\tilde u_i}| + |{\bf{v}}_i^T{\bf{x}} + {\tilde v_i}| \le \delta ({\bf{w}}_i^T{\bf{x}} + {\tilde \omega _i})\]
which can be expressed as the following linear inequalities
\[\begin{array}{l}
~~({\bf{u}}_i^T + {\bf{v}}_i^T){\bf{x}} + {{\tilde u}_i} + {{\tilde v}_i} \le \delta ({\bf{w}}_i^T{\bf{x}} + {{\tilde \omega }_i})\\
~~({\bf{u}}_i^T - {\bf{v}}_i^T){\bf{x}} + {{\tilde u}_i} - {{\tilde v}_i} \le \delta ({\bf{w}}_i^T{\bf{x}} + {{\tilde \omega }_i})\\
~~({\bf{v}}_i^T - {\bf{u}}_i^T){\bf{x}} - {{\tilde u}_i} + {{\tilde v}_i} \le \delta ({\bf{w}}_i^T{\bf{x}} + {{\tilde \omega }_i})\\
 - ({\bf{u}}_i^T + {\bf{v}}_i^T){\bf{x}} - {{\tilde u}_i} - {{\tilde v}_i} \le \delta ({\bf{w}}_i^T{\bf{x}} + {{\tilde \omega }_i}).
\end{array}\]
Meanwhile, in practical applications, a depth constraint can be additionally considered as
\[{d_{\min }} \le {\bf{w}}_i^T{\bf{x}} + {\tilde \omega _i} \le {d_{\max }}\]
where ${d_{\min }}$ and ${d_{\max }}$ denote the minimal and maximal depth, respectively.
It can be converted into two linear constraints
\[ - {\bf{w}}_i^T{\bf{x}} - {\tilde \omega _i} \le  - {d_{\min }}~~~~{\mathrm{and}}~~~~{\bf{w}}_i^T{\bf{x}} + {\tilde \omega _i} \le {d_{\max }}\]
In this case, $\kappa = 6$ in the constraint of (\ref{I-5}).

For the ${\ell_\infty }$ norm, i.e., $p=\infty $, the constraint (\ref{I-8}) becomes
\[\max \left\{ {|{\bf{u}}_i^T{\bf{x}} + {{\tilde u}_i}|,|{\bf{v}}_i^T{\bf{x}} + {{\tilde v}_i}|} \right\} \le \varepsilon ({\bf{w}}_i^T{\bf{x}} + {\tilde \omega _i})\]
which is equivalent to
\[|{\bf{u}}_i^T{\bf{x}} + {\tilde u_i}| \le \delta ({\bf{w}}_i^T{\bf{x}} + {\tilde \omega _i})~~{\rm{and}}~~|{\bf{v}}_i^T{\bf{x}} + {\tilde v_i}| \le \delta ({\bf{w}}_i^T{\bf{x}} + {\tilde \omega _i}).\]
Using (\ref{I-4}), equivalent linear inequalities of the constraint (\ref{I-8}) can be derived as follows
\[{\bf{u}}_i^T{\bf{x}} + {\tilde u_i} \le \delta ({\bf{w}}_i^T{\bf{x}} + {\tilde \omega _i}),~- {\bf{u}}_i^T{\bf{x}} - {\tilde u_i} \le \delta ({\bf{w}}_i^T{\bf{x}} + {\tilde \omega _i})\]
\[{\bf{v}}_i^T{\bf{x}} + {\tilde v_i} \le \delta ({\bf{w}}_i^T{\bf{x}} + {\tilde \omega _i}),~- {\bf{v}}_i^T{\bf{x}} - {\tilde v_i} \le \delta ({\bf{w}}_i^T{\bf{x}} + {\tilde \omega _i}).\]
The effectiveness of the ${\ell_\infty }$ norm has been demonstrated in various problems in computer vision [9]--[11].

\subsection{Connection to Existing Works}

A reformulation of (1) has been considered in [5] as
\begin{equation}\label{I-9}
\begin{split}
&\mathop {{\rm{minimize}}}\limits_{{\bf{x}},{\bf{z}}} {\rm{ }}\sum\limits_i {{z_i}} \\
\textrm{subject to  } ~&\left| {{\bf{a}}_i^T{\bf{x}} - {y_i}} \right| \le \delta  + {z_i}L,~~{z_i} \in \{ 0,1\}
\end{split}
\end{equation}
where $L$ is a large positive constant. For a solution ${{\bf{z}}^*}$ of (\ref{I-9}), it is easy to see that ${\rm{supp}}({{\bf{z}}^*}) = {\rm{supp}}({{\bf{s}}^*})$, which implies the equivalence of the formulations (2), (3), (5) and (\ref{I-9}) for solving the maximum consensus problem (1). To solve the maximum consensus problem exactly and efficiently, a guaranteed outlier removal (GORE) approach based on mixed integer linear programming has been proposed in [5] to reduce the runtime of exact algorithms. But it does not scale to high-dimensional problems, e.g., large-scale multiview reconstruction.

Very recently, deterministic approximate methods have been proposed in [4], [24].
These methods reformulate the consensus maximization problem with linear complementarity constraints,
and employ the Frank-Wolfe optimization scheme and alternating direction method of multipliers (ADMM)
to efficiently solve the reformulations.
These algorithms are efficient and effective for low-dimensional problems such as fundamental matrix and homography estimation,
but they still do not scale to large-scale multiview reconstruction problems.

It is popular to solve convex relaxed formulations of (1), e.g., $\ell_1$ approximation [6], [14], [15].
The most efficient $\ell_1$ method [6] considers a formulation as
\begin{equation}\label{I-10}
\begin{split}
&\mathop {{\rm{minimize }}}\limits_{{\bf{x}},{\bf{s}}} {{\bf{1}}^T}\tilde{\bf{ s}} \\
\textrm{subject to  }~&{\bf{Ax}} \le {\bf{b}} + \tilde{\bf{ s}},~~\tilde{\bf{ s}} \ge 0.
\end{split}
\end{equation}
This $\ell_1$ minimization formulation can be viewed as a convex relaxation of (5), where the nonconvex $\ell_0$ norm is replaced by its convex envelope, i.e., the $\ell_1$ norm. Meanwhile, the structure in ${\bf{s}} \otimes {{\bf{1}}_{\kappa  \times 1}}$ in the constraint is ignored and, hence, the number of unknown parameters in $\tilde{\bf{ s}}$ is $\kappa M$ while that in ${\bf{s}}$ is $M$. Since in our formulation the dimension of variables is significantly reduced, it can be solved more efficiently with lower computational complexity (see section IV).
The dimension reduction does not only reduce the computational complexity, but also yields improved results.

The $\ell_1$ relaxation is convenient due to its convexity and that well-developed LP solvers can be directly applied. However, the convex relaxation may degrade the performance. It has been demonstrated in the sparse recovery researches that, the $\ell_0$ or $\ell_q$ ($0 < q < 1$) norm can usually yield a sparser solution than the $\ell_1$ norm [12]. Empirical results have shown that the $\ell_1$ minimization (10) is likely to remove inliers in some conditions [6], [24]. Since the $\ell_0$ and $\ell_q$ norm penalties tend to yield a sparser solution, they can be expected to reduce the possibility of removing true inliers. In this regard, a re-weighted $\ell_1$ method has been proposed recently in [29].

\section{Proposed Algorithms}

Generally, it is difficult to directly solve the nonconvex $\ell_0$ minimization problem (5). In this section, we first propose an algorithm to solve a convex relaxed version of it. Then, we develop an iteratively reweighted algorithm to approximately solve the ideal $\ell_0$ minimization problem (5).

\subsection{$\ell_1$ Algorithm with Reduced Dimension}
We consider a convex relaxation of (5) via replacing the $\ell_0$-norm by its convex envelope, the $\ell_1$-norm, as
\begin{equation}\label{II-11}
\begin{split}
&\mathop {{\rm{minimize }}}\limits_{{\bf{x}},{\bf{s}}} {\left\| {\bf{s}} \right\|_1}\\
\textrm{subject to  } ~&{\bf{Ax}} \le {\bf{b}} + {\bf{s}} \otimes {{\bf{1}}_{\kappa  \times 1}},~~{\bf{s}} \ge 0.
\end{split}
\end{equation}
This formulation is similar to that considered in [6],
except for that the number of slack variables in (11) is $M$
while that in the $\ell_1$ algorithm [6] is $\kappa M$.
The dimension reduction would result in a speedup.
The dual problem of (11) is given by
\begin{equation}\label{II-12}
\begin{split}
&\mathop {{\rm{max }}}\limits_{{\bf{y}},{\bf{v}}}  - {{\bf{b}}^T}{\bf{y}}\\
\textrm{subject to  } ~&{{\bf{A}}^T}{\bf{y}} = {\bf{0}}\\
&{{\bf{1}}_{M \times 1}}-{\bf{Jy}}-{\bf{v}}={\bf{0}}\\
&{\bf{y}} \ge 0,~~{\bf{v}} \ge 0
\end{split}
\end{equation}
where ${\bf{y}} \in {\mathbb{R}^{\kappa M}}$ and ${\bf{v}} \in {\mathbb{R}^M}$ are the dual variables, and ${\bf{J}} = {{\bf{I}}_M} \otimes {{\bf{1}}_{1 \times \kappa }} \in {\mathbb{R}^{M \times (\kappa M)}}$. ${{\bf{I}}_M}$ is an identity matrix of size $M$.
The dual problem (12) can be solved by well-developed LP solvers. The algorithm is summarized as follows.

\begin{table}[!h]
\begin{tabular}{p{8.4cm}}
\toprule
\textbf{Algorithm 1:} $\ell_1$ algorithm with reduced dimension\\
\midrule
\hangafter 1
\hangindent 2em
\noindent
\textbf{Input:} The set of measurements $\Omega$, inlier threshold $\delta > 0$.\\
\textbf{Begin:} \\
~~~Construct ${\bf{A}}$ and ${\bf{b}}$ from the measurements $\Omega$.\\
~~~Solve the LP problem (12) to obtain ${\bf{s}}$.\\
~~~Remove the residuals for which ${s_i}>0$.\\
\textbf{End} \\
\hangafter 1
\hangindent 2em
\noindent
\textbf{Output:} A subset ${I} \subseteq \Omega$ of the measurements for which $\exists {\bf{x}}$ such that ${E_i}({\bf{x}}) \le \delta $, $\forall i \in {I}$.\\
\bottomrule
\end{tabular}
\end{table}

\subsection{Iteratively Reweighted Algorithm}

As shown in [6], [24], the $\ell_1$ method probably removes true inliers in practical applications. This is explained in the last section, as the ${\ell _1}$ minimization may yield a solution not sparse enough. Ideally, we should minimize the number of nonzero elements in ${\bf{s}}$, i.e., the ${\ell _0}$ minimization problem (5). However, exact solving of (5) is difficult. Inspired by the success of iteratively reweighted methods in sparse recovery researches [13], we propose an iteratively reweighted algorithm to approximately solve the intractable problem (5).

First, we approximate the ${\ell _0}$ norm by the ${\ell _q}$ norm
with a small value of $q$ (e.g., $q=0.1$ in the experiments).
The ${\ell _q}$ norm usually yields a sparser solution than ${\ell _1}$ norm [20], [21].
Then, at the $(k+1)$-th iteration, the ${\ell _q}$ norm is
approximated via first-order expansion (linearization) at ${\bf{s}}^k$ obtained at the $k$-th iteration as
\begin{equation}
\left\| {{{\bf{s}}}} \right\|_{q,\varepsilon }^q \approx \sum\limits_{i = 1}^M {{{\left( {\left| {s_i^k} \right| + \varepsilon } \right)}^{q - 1}}\left| {{s_i}} \right|}\notag
\end{equation}
where $\varepsilon $ is a small positive constant. Let
\begin{equation}\label{II-13}
{{\bf{w}}^{k + 1}} = \left[ {{{\left( {\left| {s_1^k} \right| + \varepsilon } \right)}^{q - 1}}, \cdots {{\left( {\left| {s_M^k} \right| + \varepsilon } \right)}^{q - 1}}} \right]
\end{equation}
denote the updated weighting vector based on ${{\bf{s}}^k}$ of the $k$-th iteration,
then, the iteratively reweighted algorithm update the parameters at the $(k+1)$-th iteration as
\begin{equation}\label{II-14}
\begin{split}
&\mathop {{\rm{minimize }}}\limits_{{\bf{x}},{\bf{s}}} {\left\| {{{\bf{w}}^{k + 1}} \odot {\bf{s}}} \right\|_1}\\
\textrm{subject to  } &~{\bf{Ax}} \le {\bf{b}} + {\bf{s}} \otimes {{\bf{1}}_{\kappa  \times 1}},~~{\bf{s}} \ge 0
\end{split}
\end{equation}
where $\odot$ denotes the Hadamard product.

Similar to (12), the problem (14) can be solved by the duality approach.
The dual problem of (14) is given by
\begin{equation}\label{II-15}
\begin{split}
&\mathop {{\rm{max }}}\limits_{{\bf{y}},{\bf{v}}}  - {{\bf{b}}^T}{\bf{y}}\\
\textrm{subject to  } ~&{{\bf{A}}^T}{\bf{y}} = {\bf{0}}\\
&{{\bf{w}}^{k + 1}} - {\bf{Jy}} - {\bf{v}} = {\bf{0}}\\
&{\bf{y}} \ge 0, ~~{\bf{v}} \ge 0.
\end{split}
\end{equation}
The iteratively reweighted algorithm is summarized as follows.

\begin{table}[!h]
\begin{tabular}{p{8.4cm}}
\toprule
\textbf{Algorithm 2:} Iteratively reweighted algorithm\\
\midrule
\hangafter 1
\hangindent 2em
\noindent
\textbf{Input:} The set of measurements $\Omega$, inlier threshold $\delta > 0$, $\varepsilon > 0$, $q \in (0,1)$, an initialization ${{\bf{w}}^1} = {{\bf{1}}_{M \times 1}}$.\\
\textbf{Begin:} \\
~~~Construct ${\bf{A}}$ and ${\bf{b}}$ from the measurements $\Omega$.\\
~~~\textbf{For} $k = 1,2, \cdots ,K$\\
~~~~~~~~Solve (15) with ${{\bf{w}}^k}$ to obtain ${{\bf{s}}^k}$.\\
~~~~~~~~Update ${{\bf{w}}^{k+1}}$ via (13) based on ${{\bf{s}}^k}$.\\
~~~\textbf{End For}\\
~~~Set ${\bf{s}} = {{\bf{s}}^K}$ and remove the residuals for which ${s_i} > 0$.
\textbf{End} \\
\hangafter 1
\hangindent 2em
\noindent
\textbf{Output:} A subset ${I} \subseteq \Omega$ of the measurements for which $\exists {\bf{x}}$ such that ${E_i}({\bf{x}}) \le \delta $, $\forall i \in {I}$.\\
\bottomrule
\end{tabular}
\end{table}

This algorithm iteratively updates the weighting vector and
solves the LP problem (\ref{II-15}) $K$ times.
In practical applications, a small value of $K$ can yield
sufficiently good performance, as shown in the experiments.

\section{Experiments}

\begin{figure}[!t]
\centering
{\includegraphics[scale = 0.82]{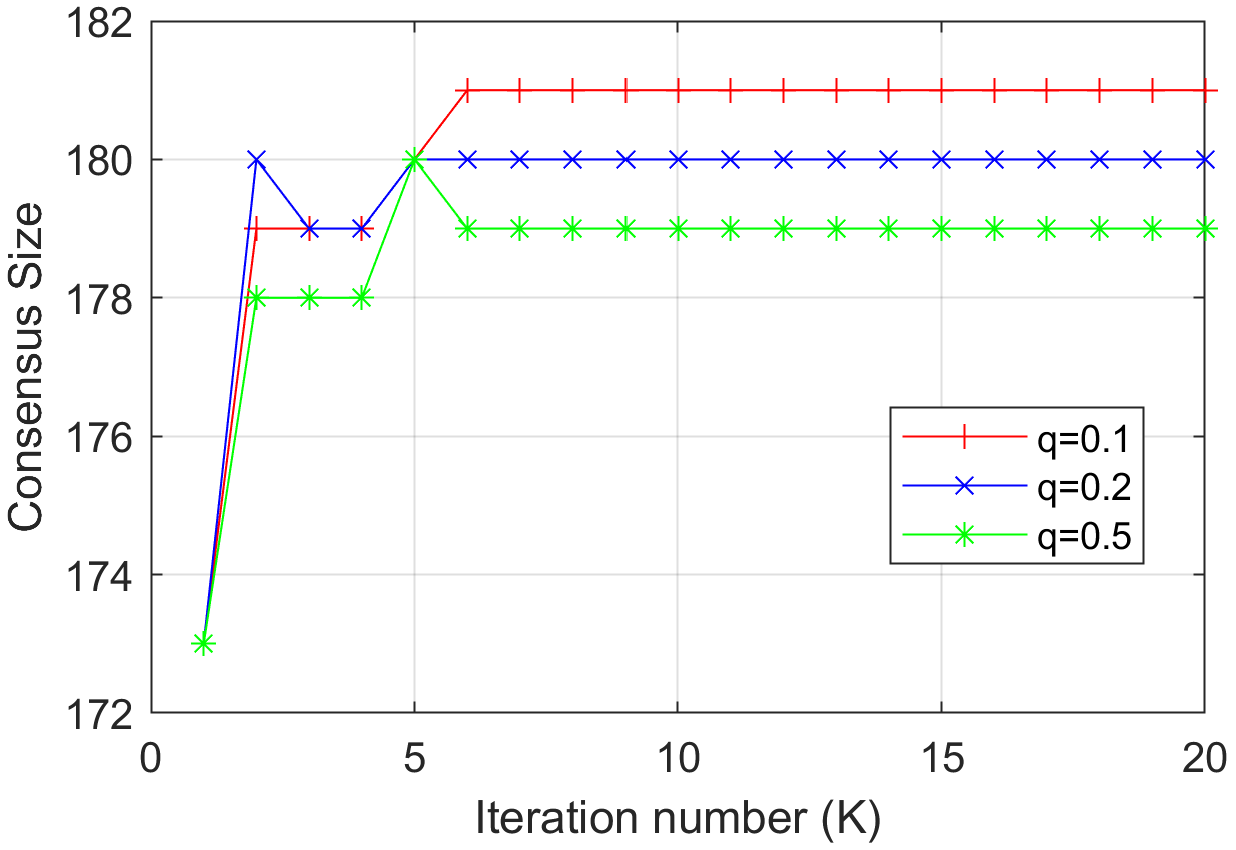}}
\caption{Results of Algorithm 2 versus iteration number for different values of $q$ (the outlier ratio is 50\%).}
\label{figure1}
\end{figure}

\begin{figure}[!t]
\centering
\subfigure[Number of remaining inliers (consensus size)]{
{\includegraphics[scale = 0.82]{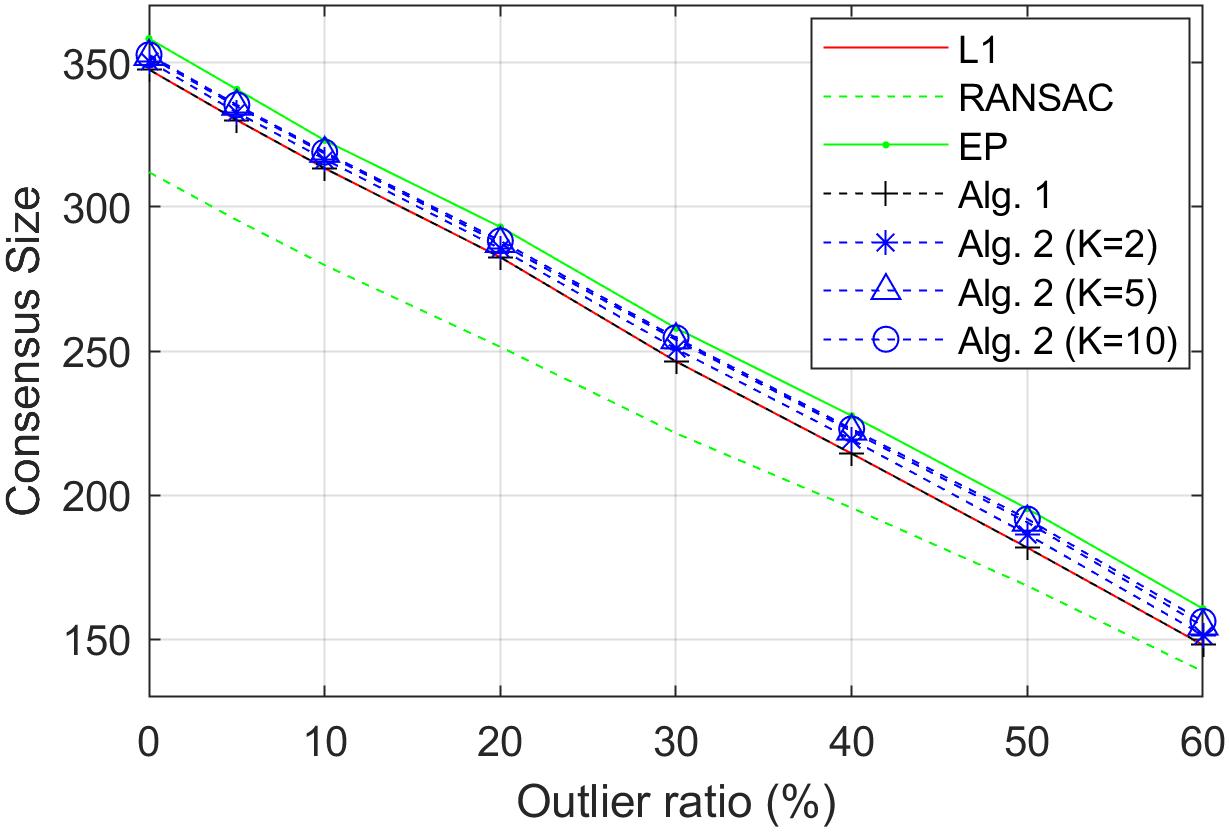}}}\\
\subfigure[Runtime]{
{\includegraphics[scale = 0.82]{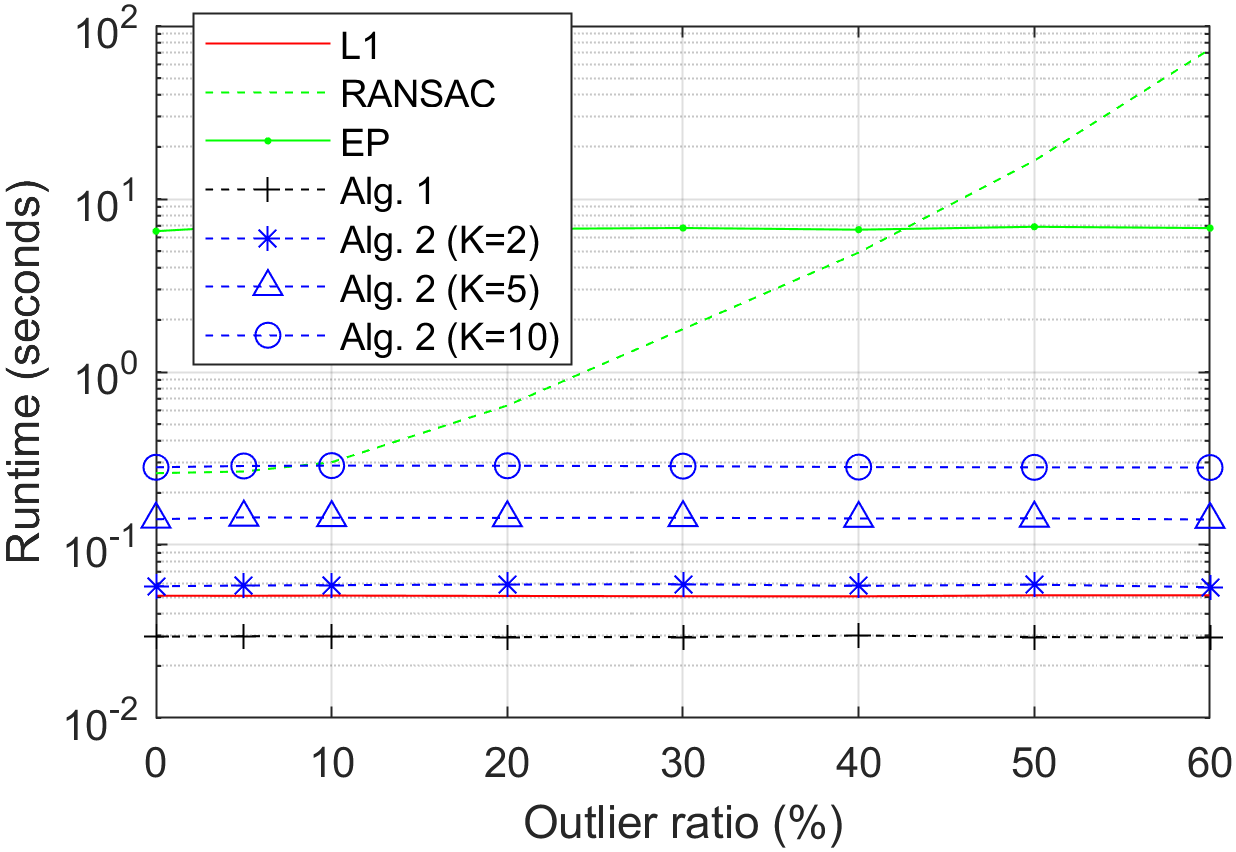}}}
\caption{Results on robust linear regression versus outlier ratio.}
\label{figure2}
\end{figure}

This section evaluates the proposed algorithms via experiments in comparison with the following methods:

\hangindent 2.2em
{(a)} RANSAC [26]: $\rho=0.99$ is used for stopping criterion.

\hangindent 2.2em
{(b)} Exact penalty (EP) method [24]\footnote{Code at: https://www.researchgate.net/publication/320707327\_demo\_pami.}:
initialized by the solution of the least squares method.
It deterministically solves a reformulation of the consensus maximization problem with linear complementarity constraints by the Frank-Wolfe algorithm.

\hangindent 2.2em
{(c)} $\ell_1$ method [6]\footnote{Code is available at: http://www.maths.lth.se/matematiklth/personal/calle/.}:
it solves the dual problem of (10) using an LP solver,
which is one of the most efficient algorithms that suitable for large-scale 3D-reconstruction.

\hangindent 2.2em
{(d)} ${\ell _\infty }$ method [11]: solved using Gugat's algorithm [22].
This method also use slack variables and minimizes the maximum slack value.
It repeatedly removes the data with the largest slack value, until the largest slack value is not greater than zero.

We use $q=0.1$ and $\varepsilon  = {10}^{-3}$ for Algorithm 2.
SeDuMi [23] is used to solve the involved LP problems in all these algorithms.
The experiments were implemented in MATLAB and run on a laptop with 2 GHz Intel I7 CPU and 16 GB RAM.

\subsection{Robust Linear Regression on Synthetic Data}

\begin{table*}[!t]
\renewcommand\arraystretch{1.2}
\caption{Reconstruction results on the house dataset. It contains 23 cameras and 35470 3D-points which are visible in at least 2 images. The RMSE without removal of any outlier is 4.76 pixels.}
\centering
\begin{tabular}{|c|c|c|c|c|c|}
\hline
& & $\ell_\infty$ (Gugat's) [11]& $\ell_1$ [6]  & Alg. 1 & Alg. 2 ($K=2$)\\
\hline
\multirow{4}{*}{House } &Removed outliers	&2396&	862	&722&	498\\
\cline{2-6}
& Remaining inliers&	99936	&101470	&101610	&101834\\
\cline{2-6}
&RMSE (pixels)&	0.9119&	0.5982	&0.6007	&0.6063\\
\cline{2-6}
&Runtime (seconds)&	50878	&975&	322	&477\\
\hline
\end{tabular}
\end{table*}

\begin{table*}[!t]
\renewcommand\arraystretch{1.2}
\caption{Reconstruction results on the cathedral dataset. It contains 17 cameras and 16961 3D-points which are visible in at least 2 images. The RMSE without removal of any outlier is 3.24 pixels.}
\centering
\begin{tabular}{|c|c|c|c|c|c|}
\hline
& & $\ell_\infty$ (Gugat's) [11]& $\ell_1$ [6]  & Alg. 1 & Alg. 2 ($K=2$)\\
\hline
\multirow{4}{*}{Cathedral }&Removed outliers	&1594	&652	&541	&428\\
\cline{2-6}
&Remaining inliers	&44451	&45393	&45504	&45617\\
\cline{2-6}
&RMSE (pixels)	&1.1343	&0.8036	&0.8075	&0.8178\\
\cline{2-6}
&Runtime (seconds)	&2354	&259	&76	&134\\
\hline
\end{tabular}
\end{table*}

\begin{table*}[!t]
\renewcommand\arraystretch{1.2}
\caption{Reconstruction results on the college dataset. It contains 57 cameras and 8990 3D-points which are visible in at least 2 images. The RMSE without removal of any outlier is 4.16 pixels.}
\centering
\begin{tabular}{|c|c|c|c|c|c|}
\hline
& & $\ell_\infty$ (Gugat's) [11]& $\ell_1$ [6]  & Alg. 1 & Alg. 2 ($K=2$)\\
\hline
\multirow{4}{*}{College }&Removed outliers	&966	&459	&319	&179\\
\cline{2-6}
&Remaining inliers	&26756	&27263	&27403	&27543\\
\cline{2-6}
&RMSE (pixels)	&1.0754	&0.5134	&0.5197	&0.5562\\
\cline{2-6}
&Runtime (seconds)	&4880	&162	&51	&79\\
\hline
\end{tabular}
\end{table*}

\begin{figure*}[!t]
\centering
\subfigure[House]{
{\includegraphics[scale = 0.8]{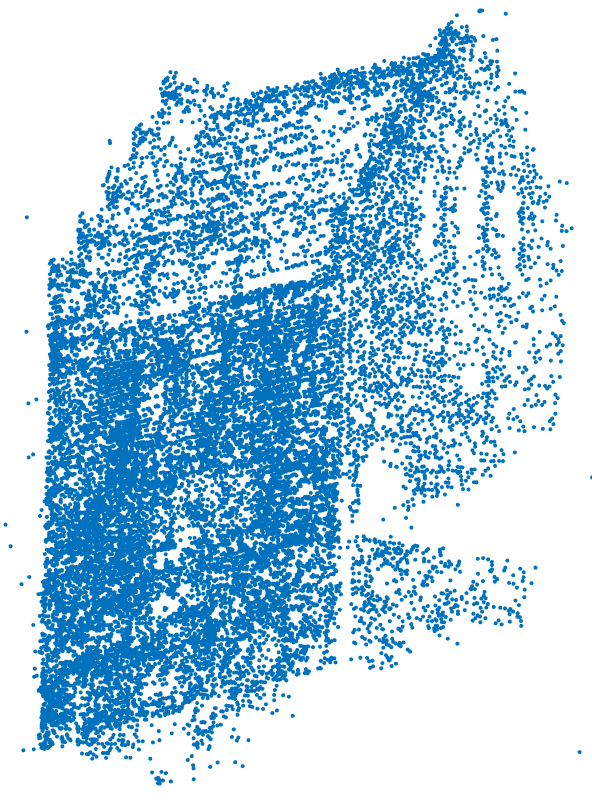}}}~~~~~~
\subfigure[Cathedral]{
{\includegraphics[scale = 0.5]{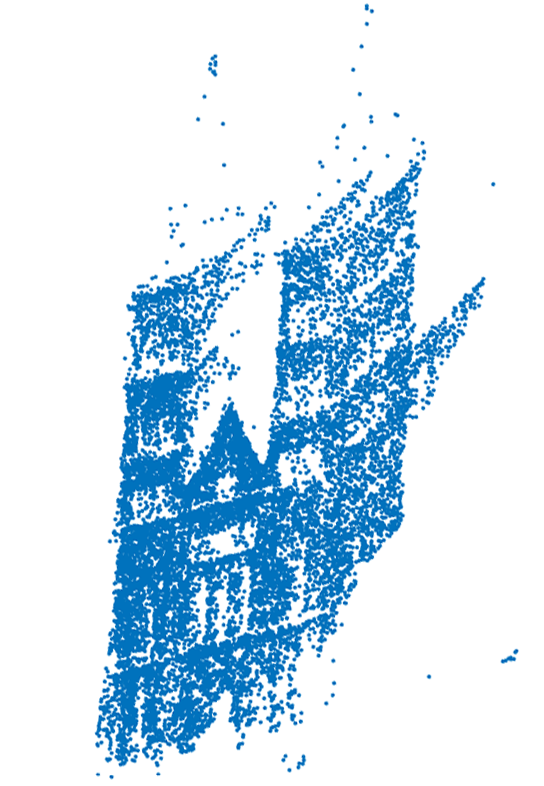}}}
\subfigure[College]{
{\includegraphics[scale = 0.85]{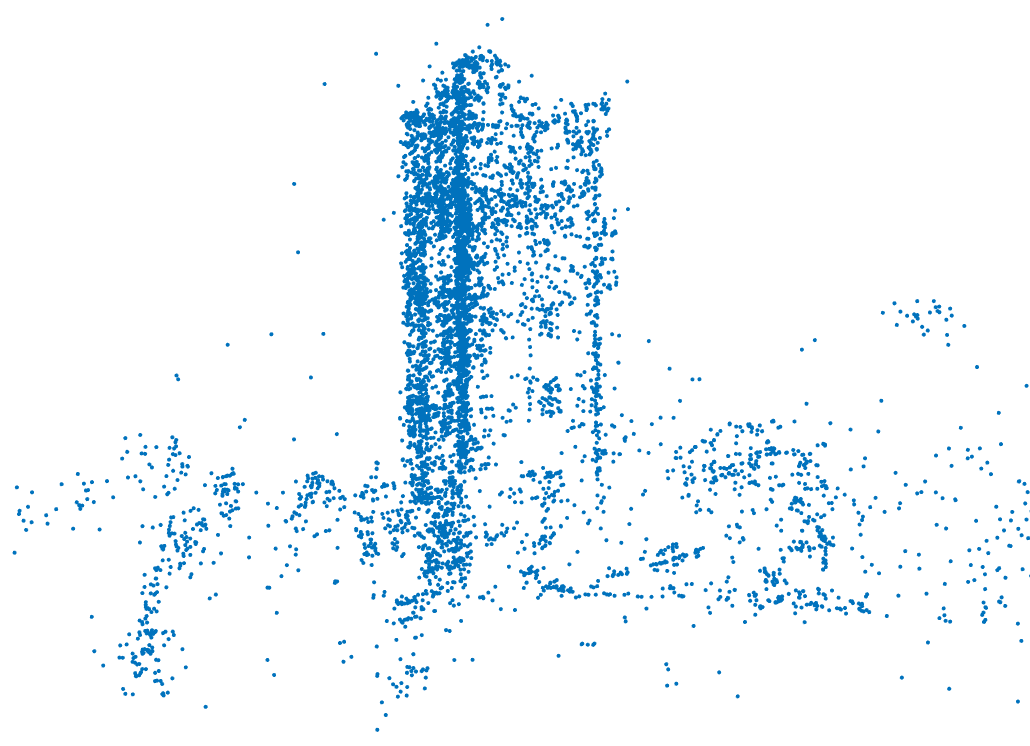}}}
\caption{The resulting reconstruction on the three datasets.}
\label{figure3}
\end{figure*}

Before proceeding to the main interest of 3D-reconstruction,
we repeat a synthetic experiment in [24] on a small linear regression problem with synthetic data. We generated $M = 500$ points ${\bf{y}} = {\bf{Ax}}$ with ${\bf{A}} \in {\mathbb{R}^{500 \times 8}}$ and ${\bf{x}} \in {\mathbb{R}^8}$. The elements of ${\bf{A}}$ follow a uniform distribution in $[-1,1]$. ${\bf{y}}$ is perturbed by white Gaussian noise with standard deviation of 0.1. To simulate outliers, a part of the elements in ${\bf{y}}$ are corrupted by much higher Gaussian noise with standard deviation of 1.

Fig. 1 shows the performance of the iteratively reweighted algorithm (Algorithm 2) versus the iteration number $K$ for different values of $q$, e.g., $q\in\{0.1,0.2,0.5\}$. It can be seen that,
Algorithm 2 converges within a few iterations, e.g., $K<10$.
Fig. 2 shows the results of the algorithms for different outlier ratio. Each result is an average over 100 independent run. Compared with the ${\ell _1}$ method, Algorithm 1 removes a same number of outliers in all cases, while Algorithm 2 removes fewer outliers. Algorithm 2 can achieve sufficiently good performance within a few iterations, e.g. $K = \{ 2,5\} $. The EP method yields the largest consensus size. In terms of runtime, Algorithm 1 is the fastest while Algorithm 2 is much faster than EP.

\subsection{Global SfM on Real World Datasets}

We consider the full 3D-reconstruction experiments in [6] and use the known-rotation formulation for outlier removal, which is a procedure of RANSAC followed by outlier removal and bundle adjustment [18]. The initial camera rotation and image data are obtained using RANSAC for pairwise images. The global outlier removal is achieved over the structure and translation of the cameras. The inlier threshold is selected corresponding to a reprojection error tolerance of 5 pixels.

Three open datasets\footnote{Available online at http://www.maths.lth.se/matematiklth/personal/calle/.}
are used, including a house (consists of 23 cameras and 29220 3D points projected into 35470 image points), a cathedral (consists of 17 cameras and 16961 3D points projected into 46045 image points), and a college (consists of 57 cameras and 8990 3D points projected into 27722 image points). In these datasets, SIFT descriptors [17] are used to generate point correspondences and, then, RANSAC is used to discard outliers and determine orientations between pairs of cameras.

Since RANSAC and EP are not suitable for large-scale problems, only the ${\ell _1 }$ [6] and ${\ell _\infty }$ [11] methods are compared here.
We use $q = 0.1$ and $K = 2$ for Algorithm 2.
Fig. 3 illustrates the resulting reconstruction on the three datasets.
Table 1 compares the reconstruction results, including removed outliers, remaining inliers, root-mean-squared-error (RMSE) of reprojection in pixels, and runtime in seconds. Without removal of any outliers, the reprojection RMSEs in reconstructing the house, cathedral and college are respectively 4.76, 3.24, and 4.16 pixels. It can be seen that each algorithm can significantly reduce the RMSE, which implies the effective removal of some outliers. A significant drop in the reprojection RMSE implies the removed data are true outliers.

Algorithm 1 is more than three times faster than the ${\ell _1 }$ method, which is due to the reduced dimension. For example, in the cathedral experiment, the number of variables in the ${\ell _1 }$ method is 327201 (${\bf{x}} \in {\mathbb{R}^{50931}}$ contains the parameters of the 3D points and camera translations, and $\tilde{\bf{ s}} \in {\mathbb{R}^{276270}}$ contains the slack variables). Whereas, in our formulation the number of variables is 96976, with ${\bf{x}} \in {\mathbb{R}^{50931}}$ and ${\bf{ s}} \in {\mathbb{R}^{46045}}$. In the experiments, $\kappa = 6$ as a depth constraint is also considered.

Compared with the $\ell_1$ method, Algorithm 1 removes fewer outliers while yields a comparable RMSE. The removal of fewer outliers is in contrast to the results in the linear regression experiment with synthetic data. Algorithm 2 removes much fewer outliers compared with the other algorithms, while being much faster than the ${\ell _1 }$ and ${\ell _\infty }$ methods.

\section{Conclusion}
In Global Structure-from-Motion, feature point matching often generates some mismatches, which gives rise to outliers. This work developed two efficient methods to detect and remove such outliers. Compared with existing methods, the new methods use a dimension reduced formulation, which significantly reduces the computational complexity. Realistic multiview reconstruction experiments demonstrated that, the new algorithms are much faster than state-of-the-art algorithms (e.g., more than three times faster than the ${\ell_1 }$ method) while give an improved solution. Due to their efficiency and effectiveness, the new methods could be useful in practical large-scale Structure-from-Motion applications.

\end{document}